\documentclass[conference]{IEEEtran}
\IEEEoverridecommandlockouts
\usepackage{cite}
\usepackage{amsmath,amssymb,amsfonts}
\usepackage{graphicx}
\usepackage{enumitem}
\usepackage[table]{xcolor}
\usepackage{multirow}
\definecolor{Gray}{gray}{0.9} % Light gray for row 
\definecolor{LightBlue}{RGB}{173, 216, 230} % Light blue
\usepackage{textcomp}
\usepackage{xcolor}
\usepackage{graphicx} % For including graphics if needed
\usepackage{booktabs} % For enhanced table layout with \toprule, \midrule, \bottomrule
\usepackage{amsmath}
\usepackage{amssymb}
\usepackage{algorithm}
\usepackage{algorithmic}
\usepackage[dvipsnames]{xcolor}
\usepackage{subcaption}
\usepackage{comment}
\usepackage{soul}
\usepackage{array} % For better array handling
\usepackage[table,xcdraw]{xcolor} % For coloring rows and cells
\usepackage{caption} % For managing captions, optional 

\def\BibTeX{{\rm B\kern-.05em{\sc i\kern-.025em b}\kern-.08em
    T\kern-.1667em\lower.7ex\hbox{E}\kern-.125emX}}

\begin{document}

\title{INTARG: Informed Real-Time Adversarial Attack Generation for Time-Series Regression}

\author{
    \IEEEauthorblockN{
        Gamze Kirman Tokgoz\textsuperscript{1}, 
        Onat Gungor\textsuperscript{2}, 
        Tajana Rosing\textsuperscript{2}, 
        Baris Aksanli\textsuperscript{1}
    }
    \IEEEauthorblockA{
        \textsuperscript{1}San Diego State University, San Diego, CA, USA \\
        \textsuperscript{2}University of California, San Diego, CA, USA \\
        \{gkirmantokgoz1518, baksanli\}@sdsu.edu, \{ogungor, tajana\}@ucsd.edu
    }
}

% \and
% \and
% \and
% \and
% \and
\newcommand{\Design}[0]{\textsc{INTARG}}

\maketitle

\begin{abstract}
Time-series forecasting aims to predict future values by modeling temporal dependencies in historical observations. It is a critical component of many real-world systems, where accurate forecasts improve operational efficiency and help mitigate uncertainty and risk. More recently, machine learning (ML), and especially deep learning (DL)-based models, have gained widespread adoption for time-series forecasting, but they remain vulnerable to adversarial attacks. However, many state-of-the-art attack methods are not directly applicable in time-series settings, where storing complete historical data or performing attacks at every time step is often impractical.
% In realistic settings, a clever, strategic attacker is unlikely to attempt to perturb every time step; instead, they would act selectively to maximize prediction degradation while avoiding detection. 
This paper proposes an adversarial attack framework for time-series forecasting under an online bounded-buffer setting, leveraging an informed and selective attack strategy.
% , rather than attacking at every time step, which would be both unrealistic and inefficient.
By selectively targeting time steps where the model exhibits high confidence and the expected prediction error is maximal, our framework produces fewer but substantially more effective attacks. Experiments show that our framework can increase the prediction error up to 2.42$\times$, while performing attacks in fewer than 10\% of time steps. 
% in a realistic power-system scenario: using the Fast Gradient Sign Method (FGSM) attack on only 9.12\% of high-confidence time steps increases the RMSE on the attacked subset from 0.0840 to 0.2525, representing approximately a 3.01× increase in error.

\end{abstract}

\begin{IEEEkeywords}
Cyber Security, Resilient Machine Learning, Adversarial Attacks, Time Series Regression
\end{IEEEkeywords}

\section{Introduction} 
Many applications rely on time-series forecasting to predict future values from historical data, particularly in safety-critical domains such as power systems \cite{tang2021adversarial}, healthcare \cite{morid2023time}, finance \cite{tang2022survey}, and communication networks \cite{ferreira2023forecasting}. To achieve strong predictive performance, machine learning (ML) and deep learning (DL)-based forecasting models have become increasingly popular, as they can capture complex nonlinear temporal dependencies and often outperform traditional methods on high-dimensional or noisy data \cite{lim2021time}. In many real-world deployments, forecasts must be generated online, with predictions produced in real time from streaming data \cite{lau2025fast}. Accurate short-term forecasts in such settings enable proactive actions, including load balancing and anomaly response, while reducing uncertainty and risk \cite{eren2024comprehensive}.

%%%%%%%%%%%%%%%%%%%%%%%%%%%%%%%%%%%%%%%%%%
\begin{figure}[!t]
  \centering
  \includegraphics[
    width=\columnwidth,
    height=0.48\textheight,
    keepaspectratio
  ]{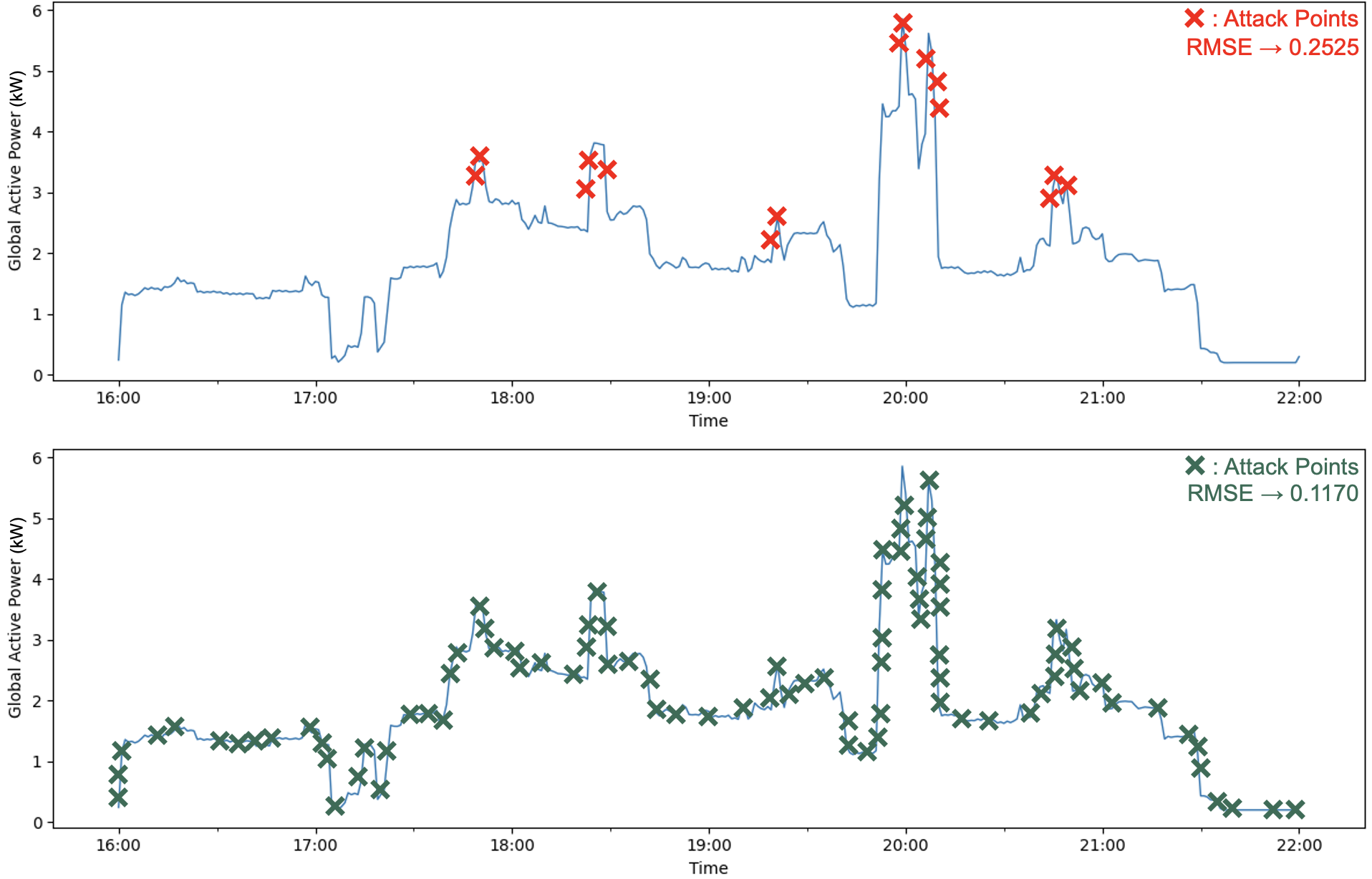}
  \caption{Power time-series (6-hour window) with informed (top) and non-selective (bottom) attacks}
  \label{fig:motivation}
\end{figure}
%%%%%%%%%%%%%%%%%%%%%%%%%%%%%%%%%%%%%%%%%%

\par Despite their strong predictive performance, DL models remain vulnerable to adversarial attacks, in which an attacker introduces small, carefully crafted perturbations to a benign input sequence to induce incorrect predictions \cite{gungor2024roldef, gungor2024rigorous, kocal2025relate}. In time-series forecasting, even minor modifications to the input history can cause large errors and compromise reliability \cite{wu2022small}. Adversarial robustness measures a model’s ability to maintain high predictive performance under such attacks, relative to clean (unperturbed) data. Most robustness studies in time-series forecasting assume access to the full historical data \cite{govindarajulu2023targeted}, an assumption that is often unrealistic. First, historical data are typically stored in secure systems with restricted access. Second, an attacker cannot wait to collect the entire history, as forecasts and downstream decisions are made in real time. In streaming deployments, adversaries must operate online as data arrive, with access limited to a recent sliding window. 
% This gap motivates the need for adversarial attack frameworks that are designed explicitly for an online bounded-buffer, streaming forecasting rather than offline settings with full access to historical data.
% \par Power consumption forecasting \cite{hong2016probabilistic} is a key power-system use case where uncertainty varies by operation regime: demand is typically predictable during routine periods but becomes more volatile during peaks and rapid ramps. This motivates using uncertainty estimations and makes power-related time-series a natural testbed for robustness in streaming forecasting.
\par Many existing adversarial attack strategies assume that the attacker perturbs the entire input sequence, which is often costly and unrealistic. In practice, an attacker may act strategically, selectively targeting high-confidence, high-impact time steps to maximize disruption. Figure \ref{fig:motivation} illustrates why informed, selective attacks are both more realistic and more damaging in time-series forecasting. It compares a selective attack strategy with a non-selective attack \cite{krishan2024adversarial} over the same 6-hour window on power consumption data. In the selective attack (top), only a small number of high-confidence, high-impact time steps are perturbed, whereas the non-selective attack (bottom) perturbs nearly every time step. Although the non-selective strategy applies more perturbations, many occur during low-impact periods, reducing the average effectiveness of each modification. By targeting critical points such as peaks or sharp ramps, the selective attacker achieves substantially greater degradation with far fewer perturbations, as reflected by the higher RMSE in the selective scenario. 
% However, in streaming settings, a fixed threshold can become miscalibrated as conditions change.
%\par In this paper, we propose a novel adversarial attack framework for streaming time-series forecasting under an online, bounded-buffer setting, as shown in Figure \ref{fig:framework}. The key idea is an informed selective attack strategy: instead of attacking at every time step, the attacker focuses on a small subset of high-confidence time steps, where confidence is defined by comparing the CQR prediction interval width $W_t$ to an adaptive threshold $T_t$ computed from recent widths (see Section III, Proposed Framework). From the attacker's perspective, this selection aims to maximize the system-level consequence per unit of perturbation by concentrating attacks on time steps where small input changes can cause the larger degradation in forecasting error. During streaming, for each new data point, a prediction interval and an adaptive, quantile-based threshold are computed, and adversarial attacks are triggered only when this confidence-based trigger is satisfied. We assess attack effectiveness by comparing prediction error before and after perturbation and by evaluating attack detection performance. Experiments on the UCI Household Power Consumption dataset \cite{individual_household_electric_power_consumption_235} demonstrate that attacking only a small fraction of high-confidence time steps can cause a large degradation in forecasting error.\\%
\par
\begin{figure*}[!t]
    \centering
    \includegraphics[width=\textwidth]{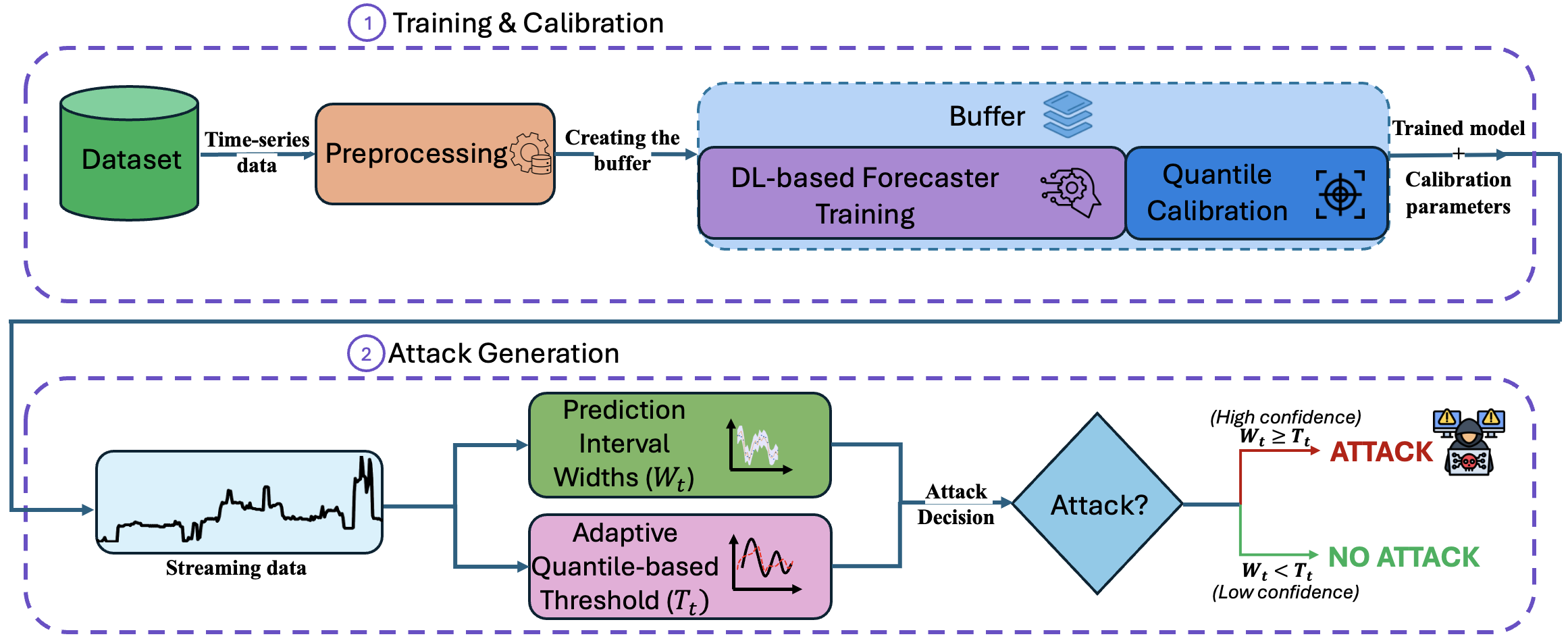}
    \caption{\Design{} components: 1) (Top) Training and Calibration and 2) (Bottom) Attack Generation}
    \label{fig:framework}
\end{figure*}

 In this paper, we propose a novel adversarial attack framework for streaming time-series forecasting under an online, bounded-buffer setting, as shown in Figure \ref{fig:framework}. \Design{} performs adversarial attacks selectively and adaptively during streaming.
The main contributions of \Design{} are as follows:
\begin{itemize}
    \item \textbf{An online bounded-buffer setting streaming attack:} %We study adversarial attacks in streaming time-series forecasting under an online bounded-buffer setting, where the model 
    We keep a fixed-size rolling buffer of recent data, reflecting real-world settings where only recent data is available.
    \item \textbf{Uncertainty-based selective attacks:} We introduce a confidence-aware selective adversarial attack strategy %that uses uncertainty information 
    to choose when to attack, perturbing only a small subset of high-confidence time steps to maximize impact under a constrained attack budget.
    \item \textbf{Adaptive thresholding method under streaming data:} We propose an adaptive quantile-based threshold computed from a rolling history of prediction interval widths to implement the selective strategy online. Unlike a fixed threshold, this adaptive mechanism can adapt to changing data conditions in streaming settings, triggering attacks only when the model is confident.
    \item \textbf{Evaluation on power-related time-series datasets:} We evaluate \Design{} on two power-related time-series forecasting applications. Our results show that the proposed method consistently degrades forecast performance, reaching up to a $2.17\times$ increase in RMSE on the Household dataset and up to a $2.42\times$ increase on the Pecan Street database over the baseline.
\end{itemize}

\section{Related Work}
%%%%%%%%%%%%%%%%%%%%%%%%%%%%%%%%%%%%%%%
%%%%%%%%%%%%%%%%%%%%%%%%%%%%%%%%%%%%%%%

\subsection{Adversarial Machine Learning}
\par Adversarial machine learning was first studied in image classification, where neural networks were shown to be vulnerable to small, carefully crafted perturbations \cite{goodfellow2015explaining, szegedyintriguing}. Subsequent work extended these vulnerabilities to time-series forecasting. Fawaz et al. \cite{fawaz2019adversarial} showed that time-series classifiers are susceptible to adversarial attacks, with perturbations exhibiting transferability and enabling black-box attacks via surrogate models.
% These findings motivate adversarial robustness for time-series forecasting, where forecasting errors can directly influence operational decisions and associated costs in real-world systems.
Several works have adapted gradient-based attacks to time-series forecasting. Costa et al. \cite{costa2024deep} extend methods such as FGSM and iterative variants like BIM to time-series inputs. Krishan et al. \cite{krishan2024adversarial} investigate untargeted white-box attacks against multivariate forecasting models, highlighting that the resulting perturbations can be visually subtle while still significantly degrading predictive performance.
\par Beyond untargeted degradation, adversaries may pursue forecasting-specific objectives, such as steering predictions or introducing systematic bias. Govindarajulu et al. \cite{govindarajulu2023targeted} investigate targeted attacks under forecasting threat models that go beyond standard classification. Dang et al. \cite{dang2020adversarial} study adversarial attacks on probabilistic autoregressive forecasters, considering objectives such as over- and under-estimation, and emphasizing the role of uncertainty and inference procedures.
% \par Energy forecasting influences operational planning and financial decisions, motivating adversarial risk studies in power contexts. Chen et al. \cite{chen2019exploiting} study black-box attacks on load forecasting, with unknown model details, while Tang et al. \cite{tang2021adversarial} examine white-box and black-box attacks on solar power forecasting and show that adversarial training can improve robustness but may worsen clean forecasting error.
\par Existing studies have also addressed challenges arising from continuously arriving data, concept drift, and computational and memory constraints. Wen et al. \cite{wen2023onenet} emphasize continual adaptation to handle drift, while Romano et al. \cite{romano2019conformalized} develop conformal methods for generating prediction intervals in real-time forecasting deployments.

\subsection{Defense and Detection Methods}

\par Defense strategies for time-series models include training-time robustness and input-level correction. Krishan et al. \cite{krishan2024adversarial} evaluate adversarial training and model hardening for FGSM/BIM attacks. In contrast, purification approaches aim to remove perturbations before inference, e.g., Chen et al. \cite{cheng2025adversarial} propose diffusion-model-based  purification for power-system event classification, by considering real-time constraints.
For adversarial input detection, Abdu et al. \cite{abdu2020detecting} use lightweight time-series descriptors with outlier detection, while Ma et al. \cite{ma2018characterizing} propose Local Intrinsic Dimensionality (LID) features in representation space to separate adversarial examples. 
% We adopt LID for attack detection; details are provided in Section IV-B2.
\par Prior work establishes that time-series forecasters are vulnerable to adversarial perturbations \cite{krishan2024adversarial}, that forecasting-specific objectives such as targeted biasing matter \cite{govindarajulu2023targeted, dang2020adversarial}, and that energy forecasting pipelines face realistic adversarial risks \cite{tang2021adversarial, chen2019exploiting}, with defenses often trading off robustness, accuracy, and computation \cite{tang2021adversarial, krishan2024adversarial, cheng2025adversarial}. Existing work does not fully reflect realistic deployment conditions where forecasting is online with a limited history window due to memory/storage constraints. Our work evaluates adversarial robustness under an online bounded-buffer setting and uses a selective attack strategy that triggers attacks only when the model is sufficiently confident, concentrating attacks on the most impactful time steps, and also achieving low detectability than the baseline method.

\section{Proposed Framework: \Design{}}
\label{sec:proposed-framework}
%\textcolor{red}{It would be helpful to outline what our idea will consist of. For example, how the attack decision is determined based on threshold selection and other factors.}
%\setlength{\parindent}{0pt}
\par Figure \ref{fig:framework} presents our adversarial attack framework for streaming time-series forecasting under an online, bounded-buffer setting. We assume that we only have access to a portion of the dataset through a fixed size bounded rolling buffer, while data arrive continuously in real-time. \Design{} consists of two stages. (1) Training \& Calibration: time-series data are preprocessed and organized into a rolling buffer. Using this buffer, we train a DL forecaster and apply quantile calibration to produce prediction intervals alongside point forecasts. (2) Attack Generation: as new samples arrive, the forecaster outputs a prediction interval to quantify model confidence. If the model confidence is higher than a threshold, we trigger an attack, leading to selective, fewer, and more impactful attacks.
% whose width $W_t$ serves as a proxy for model confidence. We compute an adaptive threshold $T_t$ from recent interval widths in the buffer and make an online decision: if $W_t \ge T_t$ (high-confidence), we trigger an attack; otherwise, we leave the input unchanged. In this setting, we focus on a small number of crafted attacks targeted at time steps where the model has high confidence -i.e., when the interval width satisfies $W_t \ge T_t$ in Figure \ref{fig:framework}, bottom. This strategy concentrates attacks on time steps where they are likely to cause the maximum possible damage in forecasting performance.

\subsection{Threat Model}
\label{sec:threat-model}
\par We study an online streaming forecasting system with a bounded-buffer where the model can observe the most recent days of historical data. We consider a white-box, test-time adversary who can determine when to perturb incoming data, but is unable to modify forecasting model’s parameters, the training data, or the anomaly detector. The attacker’s goal is to maximize the forecasting error while keeping perturbations small and bounded. To better reflect realistic settings, we assume the attacker applies perturbations selectively to only a fraction of time steps. We denote this fraction as the attack rate $\alpha \in (0,1)$, i.e., the proportion of input data that can be modified to maximize the adversarial attack impact.

\subsection{Training \& Calibration}
\label{sec:training-calibration}
\subsubsection{Preprocessing}
\label{secpreprocessing}
\par This step cleans the dataset by first replacing missing-value markers (e.g., “?”) with NaN, then converting features to numeric and dropping any rows containing NaNs. We then apply 0-1 normalization using the minimum and maximum from the initial buffer before being fed into the models. We construct a fixed-size rolling buffer representing the limited-memory of the system. This buffer is split into two parts (Figure \ref{fig:framework}, top): a training segment (75\%) and a calibration segment (25\%). 

\subsubsection{DL-based Forecaster Training}
\label{sec:forecaster}
The forecasting component uses a Convolutional Neural Network (CNN) operating on a fixed-size rolling window of recent observations. We implement a 1D-CNN (Conv1D + pooling + dropout) followed by dense layers for one-step ahead forecasting. We use a CNN because convolutional models are effective and widely adopted for time-series data, as they capture local temporal patterns efficiently \cite{wang2017time}.
 % Wang et al. \cite{wang2017time} highlight this advantage of CNN-based time series modeling. 
 Training examples are constructed by a sliding window over the training segment: a short history of recent observations as input and the next time-step value as the prediction target. The CNN is trained on this segment to obtain a clean forecasting model, which serves as a baseline for subsequent evaluations. We later compare performance under adversarial attacks to quantify the degradation in prediction error. %Once trained, the CNN is kept fixed during the streaming phase. 
 \par After offline training on the initial buffer, the model is deployed in the attack generation phase (Figure \ref{fig:framework}, bottom). As new measurements arrive, the rolling buffer is updated by discarding the oldest value and appending the newest one, and the most recent window is fed to the CNN to produce a one-step-ahead forecast $\hat y_{t+1}$, while the CQR module uses this window to form a prediction interval (see Section \ref{sec:calibration}).
\subsubsection{Quantile Calibration}
\label{sec:calibration}
\par A point forecast can be inadequate in many real-world applications where we also need a measure of model confidence. We adopt conformalized quantile regression (CQR) \cite{romano2019conformalized} to obtain calibrated prediction intervals without modifying the base forecaster. The calibration segment is used to meet the desired coverage level. \\
%In our setting, confidence is crucial because we want to launch adversarial attacks only at those time steps where the forecaster is most confident, so that small perturbations can cause the largest possible damage.
$\triangleright$ \textbf{Quantile Regression:} Quantile regression estimates conditional quantiles rather than the mean, enabling model to predict lower and upper quantiles, which can be used to create a prediction interval. When the noise is heteroscedastic (i.e., prediction error variance changes over time), quantile regression is particularly suitable. In power consumption forecasting, demand can be stable during routine periods but more unpredictable during sudden load changes. Quantile-based intervals can widen in volatile periods and narrow in stable conditions. Let $\alpha \in (0,1)$ denote the desired miscoverage level. We fit lower and upper quantiles $\alpha_{\mathrm{lo}} = \alpha/2$ and $\alpha_{\mathrm{hi}} = 1 - \alpha/2$. With $q_{\alpha_{\mathrm{lo}}}(x)$ and $q_{\alpha_{\mathrm{hi}}}(x)$, we obtain an estimated interval:
\begin{equation}
    C_{\text{QR}}(x)
    = \big[q_{\alpha_{\mathrm{lo}}}(x),\; q_{\alpha_{\mathrm{hi}}}(x)\,\big].
    \label{eq:qr_interval}
\end{equation}
\par Ideally this interval would satisfy approximately $1-\alpha$ coverage and adapt to local uncertainty: wider in noisy regions and narrower where the relationship is more stable. In practice, quantiles are estimated from finite data, so the resulting intervals may fail to achieve the desired coverage, especially for flexible models (e.g., neural networks).\\
$\triangleright$ {\textbf{Split Conformalized Quantile Regression (CQR):}} Conformalized Quantile Regression (CQR) calibrates a model's quantile interval using a presented calibration set to achieve marginal coverage at least $1-\alpha$. Specifically, it measures how often calibration points fall outside the predicted bounds and applies a correction so that the final interval achieves the desired coverage level. We use the calibrated interval width as a confidence signal to trigger selective attacks where well-calibrated intervals make this signal more reliable over time.
\par In our framework, we split the initial buffer into a training set $\mathcal{I}_1$ to fit the quantile forecaster and a calibration set $\mathcal{I}_2$ to compute conformity scores and the calibration correction for conformalization. Using the quantile predictions, we form the initial prediciton interval $\hat C(x) = \big[\hat q_{\alpha_{\mathrm{lo}}}(x),\; \hat q_{\alpha_{\mathrm{hi}}}(x)\,\big]$ and calibrate it on a set $\mathcal{I}_2$. Conformity scores (with input $X_i$ and the true value $Y_i$) are computed as:
\begin{equation}
    E_i = \max\big\{ \hat q_{\alpha_{\mathrm{lo}}}(X_i) - Y_i,\; Y_i - \hat q_{\alpha_{\mathrm{hi}}}(X_i) \big\},
    i \in \mathcal{I}_2.
\end{equation}

\par For a miscoverage level $\alpha$, we take the $(1-\alpha)\left(1+1/|I_2|\right)$-quantile of the empirical distribution of the conformity scores on the calibration set and expand the plug-in interval by this amount on both sides. The conformalized interval for x is:

\begin{equation}
\begin{split}
C_{\text{CQR}}(x)
= \bigl[\,
    \hat q_{\alpha_{\mathrm{lo}}}(x) - Q_{1-\alpha}(E, \mathcal{I}_2), \\
    \hat q_{\alpha_{\mathrm{hi}}}(x) + Q_{1-\alpha}(E, \mathcal{I}_2)
  \,\bigr].
\end{split}
\end{equation}

%\par Under exchangeability, Romano et al. \cite{romano2019conformalized} show that this interval satisfies the finite-sample coverage at least $1-\alpha$.

\par During streaming, the rolling buffer is updated at each new time step, the CNN produces a one-step-ahead forecast, and the CQR module outputs a conformalized prediction interval $C_{\text{CQR}}(x_t)$ using $\hat q_{\alpha_{\mathrm{lo}}}$,
$\hat q_{\alpha_{\mathrm{hi}}}$, and $Q_{1-\alpha}(E, \mathcal{I}_2)$ have been learned from the initial buffer. The width of this interval serves as our uncertainty measure and is later used by the adaptive thresholding mechanism in Section \ref{sec:adaptive-threshold} to determine whether the current time step should be attacked. After selecting which time steps to perturb, the next section describes how we generate adversarial perturbations.

\subsection{Attack Generation}
\label{sec:attack-generation}
\subsubsection{Prediction Interval Width}
\label{sec:interval-width}
%\hat{q}^{\mathrm{lo}}_{\alpha}(x_t)]$ --> old version
%$[\hat{q}_{\alpha_{\mathrm{hi}}}(x_t) --> new version
The CQR module outputs a calibrated prediction interval $[\hat{q}_{\alpha_{\mathrm{lo}}}(x_t), \hat{q}_{\alpha_{\mathrm{hi}}}(x_t)]$ at each time step $t$. We summarize its uncertainty using the prediction interval width $W_t$, defined as the difference between upper and lower bounds. $W_t$ captures how confident the model is about the next-step prediction: larger widths indicate higher confidence, while smaller widths indicate lower confidence. In \Design{}, $W_t$ serves as a confidence signal that is later used by the adaptive threshold module to decide whether to trigger an attack at time $t$.

\subsubsection{Adaptive Quantile-based Threshold}
\label{sec:adaptive-threshold}
We propose an adaptive quantile-based thresholding mechanism based on the recent history of quantile-prediction interval widths to decide when to attack. A fixed, manually chosen threshold would not be robust in our streaming, bounded-buffer setting, where the distribution of the data and the model’s uncertainty can change over time. Instead, by computing a rolling quantile of the interval widths, we automatically adjust the threshold to the current level of uncertainty, approximately maintain a desired attack rate, and always focus the attacks on the time steps that the model currently appears most confident about.
Because the values of $W_t$ can change over time, at time $t$ we maintain a rolling history $H_t$ consisting of the interval widths of the most recent $M$ time steps to compute the threshold. Given a target attack rate of
$\alpha \in (0,1)$ (e.g., $\alpha = 0.10$ for a $10\%$ attack rate), we compute an adaptive threshold $T_t$ at each time step as the empirical $(1-\alpha)$-quantile of the rolling history of interval widths $H_t$.
The adaptive threshold $T_t$ tracks the current distribution of interval widths. Based on this threshold, the attack is triggered at time $t$ whenever the current interval width $W_t$ exceeds (or equals) $T_t$ i.e., when $W_t \ge T_t$. This condition defines that the current prediction interval is wider than what is typical recently. Therefore, we use $W_t \ge T_t$ as the attack trigger to select only these flagged time steps. 
\subsubsection{Attack Decision}
\label{sec:attack-decision}
In many adversarial robustness studies, the attacker is assumed to perturb every data point in the test set. However, this assumption is unrealistic for real-world scenarios, where the attacker typically has limited information and does not have access to the entire dataset, especially in real-time settings. In such cases, there is no opportunity to wait until all data have been collected; instead, attacks must be launched online, as the data arrive, to maximize their impact \cite{gong2019real}. In our setting, we focus on attacks that target only selected time steps where the model is most confident, rather than perturbing every data point. We combine standard gradient-based attacks with a \emph{selective attack} policy driven by the uncertainty measure introduced in Section \ref{sec:calibration} and the adaptive threshold mechanism in Section \ref{sec:adaptive-threshold}.\\
$\triangleright$ \textbf{Gradient-Based Attacks:} We select Fast Gradient Sign Method (FGSM) \cite{goodfellow2015explaining}, Basic Iterative Method (BIM) \cite{kurakin2018adversarial}, and  Nesterov Iterative
Fast Gradient Sign Method (NI-FGSM) \cite{lin2019nesterov} in \Design{} because they are standard, widely used first-order (gradient-based) baselines for evaluating adversarial robustness. These three attacks provide complementary views of our framework: FGSM represents a simple, low-cost perturbation, BIM represents a stronger iterative attack, and NI-FGSM further strengthens iterative FGSM by adding momentum.
%We use two standard gradient-based attacks in our framework.\\
%\textit{Fast Gradient Sign Method (FGSM):} FGSM is a single step attack that perturbs the input in the direction of the sign of the loss gradient, originally proposed by Goodfellow et al. \cite{goodfellow2015explaining} for neural networks on image data.\\
%\textit{Basic Iterative Method (BIM):} BIM, also known as Iterative FGSM, extends FGSM by taking multiple small FGSM-like steps with a step size $\alpha$ and a small number of iterations, as in \cite{kurakin2018adversarial}, typically yields a stronger attack than FGSM at the cost of higher computation.\\
%\textit{Nesterov Iterative
%Fast Gradient Sign Method (NI-FGSM):} NI-FGSM extends the iterative FGSM by adding Nesterov-style accelerated momentum, computing gradients at a look-ahead point to improve transferability and strengthen the attack \cite{lin2019nesterov}.

\noindent$\triangleright$ \textbf{Attack Strategy:} We use a selective attack strategy based on the conformalized prediction intervals and the adaptive quantile-based threshold. For each time step, the CQR module produces a prediction interval and its width is used as an uncertainty measure. We maintain a rolling history of these widths and compute an adaptive threshold as a high quantile of this history. If the current width exceeds this threshold, the time step is selected and an adversarial perturbation (FGSM, BIM or NI-FGSM) is applied; otherwise, the input is left clean.

\section{Experimental Analysis}
%\textcolor{red}{This section should clearly show the state-of-the-art, evaluation metrics, datasets, and results with insights.}
%\setlength{\parindent}{0pt}

\begin{comment}
%SECOND TABLE
\begin{table*}[ht]
\centering
\caption{Attack Detection Results Comparison with Baseline}
\label{tab:detection-results}
\renewcommand{\arraystretch}{1.1}
\begin{tabular}{|c|c|c|c|c|c|c|c|c|c|}
\hline
\multirow{2}{*}{Attack Type} & \multirow{2}{*}{$\epsilon$} &
\multicolumn{2}{c|}{Attack Det. Acc.} & \multicolumn{2}{c|}{Precision} & \multicolumn{2}{c|}{Recall} & \multicolumn{2}{c|}{F1-score} \\
\cline{3-10}
 & & Ours & Baseline & Ours & Baseline & Ours & Baseline & Ours & Baseline \\
\hline
\multirow{3}{*}{FGSM} & 0.05 & 52.7\% & 30.94\% & 0.098 & 0.447 & 0.527 & 0.309 & 0.165 & 0.369\\
                      & 0.1 & 54.18\% & 24.46\% & 0.134 & 0.420 & 0.542 & 0.245 & 0.215 & 0.338\\
                      & 0.15 & 46.92\% & 23.02\% & 0.105 & 0.337 & 0.469 & 0.230 & 0.172 & 0.274\\
\hline
\multirow{3}{*}{BIM}  & 0.05 & 48.85\% & 38.85\% & 0.166 & 0.491 & 0.488 & 0.388 & 0.248 & 0.434\\
                      & 0.1 & 48.87\% & 34.53\% & 0.140 & 0.495 & 0.489 & 0.345 & 0.217 & 0.407\\
                      & 0.15 & 65.33\% & 55.40\% & 0.124 & 0.558 & 0.653 & 0.554 & 0.209 & 0.556\\
\hline
\end{tabular}
\end{table*}
\end{comment}

\subsection{Setup}
\label{sec:setup}
\subsubsection{Dataset Description}

We use two power-consumption datasets in our experiments. \textbf{(i) Individual Household Electric Power Consumption Dataset (UCI) \cite{individual_household_electric_power_consumption_235}}: This dataset contains $2,075,259$ minute-level data points with $9$ recorded features, including household electricity usage, voltage, current intensity, and active/reactive power. It provides a large-scale, computationally significant testbed. \textbf{(ii) Pecan Street Dataport \cite{ pecanstreet_dataport}}: This database contains residential electricity time-series measurements. It includes data from 25 houses in the Austin, TX region. Each house represents a separate, independent dataset, with around $500,000$ minute-level data points and $79$ features. We report results averaged over these 25 house datasets. Both datasets represent real-world power forecasting and address gaps in adversarial robustness for online settings. Their multivariate time-series structure supports sequential prediction and streaming deployment.

\subsubsection{Evaluation Metrics} 

%%%%%%%%%%%%%%%%%%%%%%%%%%%%%%%%%%%%%%%%%%%%%%%%%%%
\begin{figure*}[]
  \centering
  \includegraphics[
    width=0.98\linewidth,
  ]{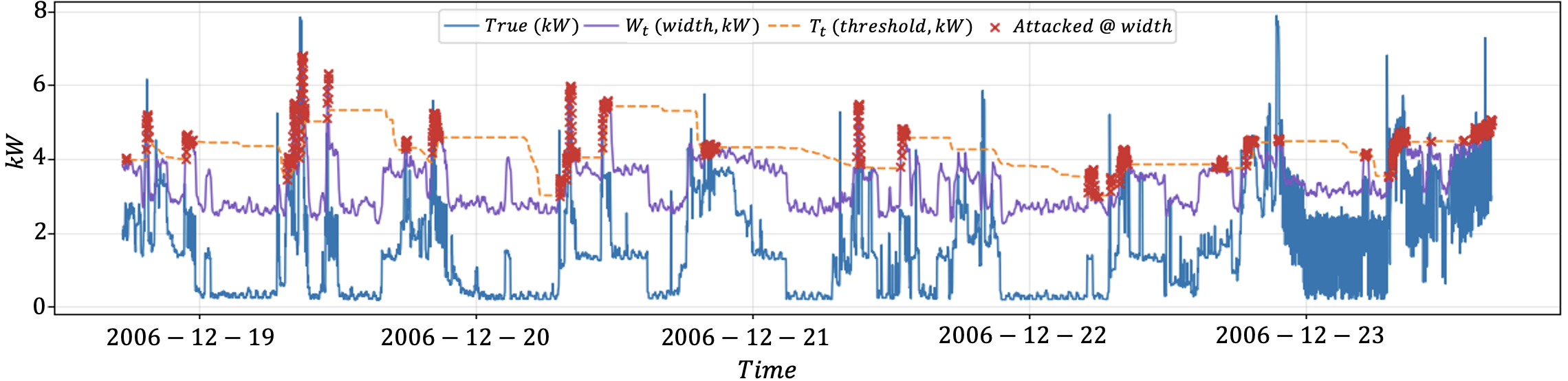}
  \caption{Our Selective Adversarial Attack Method with Adaptive Threshold}
  \label{fig:threshold}
\end{figure*}
%%%%%%%%%%%%%%%%%%%%%%%%%%%%%%%%%%%%%%%%%%%%%%%%%%%
We evaluate forecasting accuracy using root mean squared error (RMSE) and report RMSE under different adversarial attacks. To quantify attack impact without diluting the results with time steps that were never perturbed, we evaluate performance only on the attacked subset. %Table \ref{tab:attack-results-tetouan} also reports the clean RMSE on the same attacked subset using unperturbed inputs. 
Let  $T_{\text{adv}}$ denote the set of time indices at which an adversarial perturbation is applied (i.e., the trigger condition is satisfied). For each $t \in T_{\text{adv}}$, we compute the adversarial prediction $\hat y^{\mathrm{adv}}_{\text{t+1}}$, and report the RMSE on this subset:
    
\begin{equation}
    \text{RMSE}_{\text{adv}} = \sqrt{\frac{1}{|T_{\text{adv}}|} \sum_{t \in T_{\text{adv}}} \bigl(y_{t+1} - \hat{y}^{\text{adv}}_{t+1}\bigr)^{2}}
    \label{eq:rmse_adv}
\end{equation}

In addition, we also report attack detectability using the F1-score of an adversarial input detector. Specifically, we adopt the Local Intrinsic Dimensionality (LID) detector \cite{ma2018characterizing}, which distinguishes adversarial inputs from normal data by characterizing the local geometric structure of samples in representation space.

\subsubsection{Baseline}
We compare our results against baseline adversarial attack strategy \cite{krishan2024adversarial} that performs non-selective attacks, i.e., they attack every time step in the test set. Since the baseline attacker operates on the full test set while our method attacks only a subset, we ensure a fair comparison by using an attack rate that is approximately identical across methods. 
 %From this perspective, our results show that with a comparable attack percentage, our method can still produce a larger increase in RMSE than the baseline’s “attack every step” approach, showing that our confidence-based selective attacks are more efficient and more realistic in stealthy adversarial settings.

\subsection{Results}
\label{sec:results}
\subsubsection{Attack Effectiveness} 
\par To illustrate how our method works over time, Figure \ref{fig:threshold} shows an example segment of the streaming data (from the Individual Household Electric Power Consumption dataset) under a BIM attack with $\varepsilon = 0.05$. For clarity, we visualize a window of $10,000$ consecutive time steps.
Figure \ref{fig:threshold} shows the time interval together with the conformalized prediction-interval width and the adaptive quantile threshold. The purple curve corresponds to the interval width $W_t$, the orange curve to the adaptive threshold $T_t$, and the red crosses mark the time steps where an attack is launched. Attacks occur precisely when the width exceeds the threshold, and the threshold itself adapts smoothly over time to changes in the uncertainty level. Many attacks occur during peak-demand periods. This is consistent with our goal of causing maximum harm as mentioned in Figure \ref{fig:motivation}: perturbing the model during peaks can have a larger operational impact. The visualization confirms that the rolling quantile mechanism selectively targets high-confidence, high-impact regions while maintaining the desired attack rate. 

%FIRST TABLE
\begin{table}[]
\centering
\caption{Attack Performance Comparison on the Household Electric Power Consumption Dataset \cite{individual_household_electric_power_consumption_235}}
\label{tab:attack-results}
\renewcommand{\arraystretch}{1.1}
\begin{tabular}{|c|c|c|c|c|c|c|c|}
\hline
\multirow{2}{*}{Attack Type} & \multirow{2}{*}{$\epsilon$} &
\multicolumn{2}{c|}{Attack RMSE $\uparrow$} & \multicolumn{2}{c|}{F1-score $\downarrow$}\\
\cline{3-6}
 & & Ours & Baseline\cite{krishan2024adversarial} & Ours & Baseline\cite{krishan2024adversarial} \\
\hline
\multirow{3}{*}{FGSM} & 0.05 & \textbf{0.253} & 0.117 & \textbf{0.165} & 0.369\\
                      & 0.10 & \textbf{0.288} & 0.174 & \textbf{0.215} & 0.338\\
                      & 0.15 & \textbf{0.297} & 0.237 &
                      \textbf{0.172} & 0.274\\
\hline
\multirow{3}{*}{BIM}  & 0.05 & \textbf{0.280} & 0.129 & \textbf{0.248} & 0.434\\
                      & 0.10 & \textbf{0.289} & 0.203 &
                      \textbf{0.217} & 0.407\\
                      & 0.15 & \textbf{0.310} & 0.287 & \textbf{0.209} & 0.556\\
\hline
\multirow{3}{*}{NI-FGSM}  & 0.05 & \textbf{0.295} & 0.262 & \textbf{0.018} & 0.581\\
                      & 0.10 & \textbf{0.353} & 0.306 & \textbf{0.036} & 0.591\\
                      & 0.15 & \textbf{0.393} & 0.348 & \textbf{0.101} & 0.596\\
\hline
\end{tabular}
\end{table}

%SECOND TABLE 
%SECOND TABLE
\begin{table}[]
\centering
\caption{Attack Performance Comparison on the Pecan Street Database \cite{ pecanstreet_dataport}}
\label{tab:attack-results-pecan-street}
\renewcommand{\arraystretch}{1.1}
\begin{tabular}{|c|c|c|c|c|c|c|c|}
\hline
\multirow{2}{*}{Attack Type} & \multirow{2}{*}{$\epsilon$} &
\multicolumn{2}{c|}{Attack RMSE $\uparrow$} & \multicolumn{2}{c|}{F1-score $\downarrow$}\\
\cline{3-6}
 & & Ours & Baseline\cite{krishan2024adversarial} & Ours & Baseline\cite{krishan2024adversarial} \\
\hline
\multirow{3}{*}{FGSM}
  & 0.05  & \textbf{0.230} & 0.098 & \textbf{0.226} & 0.279\\
  & 0.10  & \textbf{0.245} & 0.130 & \textbf{0.228} & 0.362\\
  & 0.15  & \textbf{0.257} & 0.161 & \textbf{0.197} & 0.420\\
\hline
\multirow{3}{*}{BIM}
  & 0.05 & \textbf{0.241} & 0.100 & \textbf{0.232} & 0.261\\
  & 0.10 & \textbf{0.268} & 0.140 & \textbf{0.261} & 0.312\\
  & 0.15 & \textbf{0.289} & 0.177 & \textbf{0.266} & 0.329\\
\hline
\multirow{3}{*}{NI-FGSM}
  & 0.05 & \textbf{0.242} & 0.104 & \textbf{0.231} & 0.244\\
  & 0.10 & \textbf{0.269} & 0.141 & \textbf{0.244} & 0.300\\
  & 0.15 & \textbf{0.290} & 0.179 & \textbf{0.252} & 0.329\\
\hline
\end{tabular}
\end{table}

Table \ref{tab:attack-results} and Table \ref{tab:attack-results-pecan-street} report adversarial attack impacts on the performance of the CNN forecaster on the Individual Household Electric Power Consumption and the Pecan Street datasets, and compare \Design{} with a recent state-of-the-art adversarial attack baseline \cite{krishan2024adversarial}. We compare our selective attack strategy, which uses the adaptive quantile-based thresholding mechanism to decide when to attack, with the baseline strategy that attacks under the same experimental conditions at a similar attack ratio, without using any uncertainty information. Results are reported for FGSM, BIM, and NI-FGSM attacks and for three perturbation magnitudes $\varepsilon \in \{0.05, 0.10, 0.15\}$. For all configurations in Table \ref{tab:attack-results} and Table \ref{tab:attack-results-pecan-street}, our selective strategy achieves a higher attack RMSE than the baseline, showing that our attacks are more effective in stealthy adversarial settings, achieving up to $2.17\times$ higher RMSE on the Household dataset and up to $2.42\times$ times higher RMSE on the Pecan Street database than the baseline \cite{krishan2024adversarial}.

\subsubsection{Attack Detection}

%Ma et al. \cite{ma2018characterizing} propose the Local Intrinsic Dimensionality (LID), which characterizes adversarial regions by estimating the intrinsic dimensionality of a sample’s local neighborhood in representation space, computed from k-nearest-neighbor distance statistics. The method can be implemented efficiently via mini batch-based neighborhood estimation and forms a detector by extracting LID features across multiple network layers (using hidden activations as the distance inputs) and training a lightweight classifier to separate adversarial examples from normal/noisy inputs.
\par Attack detection evaluates how well the detector can distinguish adversarial (attacked) samples from clean (unattacked) samples. Table \ref{tab:attack-results} and Table \ref{tab:attack-results-pecan-street} report attack detection performance (F1-score) under the same streaming setting as our attack evaluation. From the attacker's point of view, a lower F1-score is desirable because it means the attacks are harder to detect, making the attack stealthier. Overall, the detector performs worse under our selective strategy, meaning it has more difficulty separating attacked samples from clean samples. For example, on the Household dataset with $\varepsilon = 0.05$, compared to the baseline method \cite{krishan2024adversarial}, the F1-score drops from $0.369$ to $0.165$ for FGSM and from $0.434$ to $0.248$ for BIM. This indicates that under our selective attack method, the detector still captures a large fraction of the true attacks, but it also misclassifies many clean samples as attacks, making its decisions less reliable. As a result, selective attacks not only increase the forecasting error, but also confuse the detection mechanism, making it harder to distinguish adversarial from normal samples.

%For each attack type and perturbation magnitude $\varepsilon \in \{0.05, 0.10, 0.15\}$, we list F1-score for our selective attacks and for the baseline method \cite{krishan2024adversarial}. We observe that our attacks make the detector less reliable, as reflected by lower F1-scores compared to the baseline across many configurations. Since the F1-score summarizes a detector’s ability to correctly distinguish attacked from clean samples, these reductions indicate degraded overall detection quality under our selective strategy.
%In all configurations, the recall of the detector under our attacks is higher than in the baseline, i.e., more true attacks are flagged. However, this comes at the cost of a much lower precision, i.e., the detector generates many more false alarms. As a consequence, the F1-score, which combines precision and recall, is consistently lower for our attacks than for the baseline. 

%\par From the defender’s point of view, this lower F1-score indicates degraded detection performance, because the detector still raises many alerts but a large fraction of them are false positives. From the attacker’s point of view, selective attacks not only increase the forecasting error, but also confuse the detection mechanism, making it harder to distinguish adversarial from normal samples.

\section{Conclusion}

\par We show that effective adversarial attacks in streaming time-series forecasting are possible under an online, bounded-buffer setting by using a selective attack strategy instead of attacking at every time step. By applying perturbations on a small subset of strategically chosen time steps, we achieve a larger degradation in forecasting performance than the existing studies under a comparable attack rate, showing that “when to attack” can matter as much as “how to attack”. In particular, our selective attacks are up to $2.17\times$ more damaging (RMSE) than the state-of-the-art baseline on the Household dataset and up to $2.42\times$ on the Pecan Street database. Our results also indicate that selective attacks can be harder to defend against, as detection performance (F1-score) is lower under our method than the baseline across both datasets. Overall, these findings support the proposed \Design{} as a practical and effective way to evaluate adversarial robustness in online time-series forecasting systems by enabling realistic selective attacks and jointly assessing their impact on forecasting degradation and detectability.

\section*{Acknowledgements}
This work has been funded in part by NSF, with award numbers \#2112665, \#2112167, \#2003279, \#2120019, \#2211386, \#2052809, \#1911095 and in part by PRISM and CoCoSys, centers in JUMP 2.0, an SRC program sponsored by DARPA.

\bibliographystyle{IEEEtran}
\bibliography{references}

@inproceedings{krishan2024adversarial,
  title={Adversarial Attacks and Defenses in Multivariate Time-Series Forecasting for Smart and Connected Infrastructures},
  author={Krishan, Pooja and Mohapatra, Rohan and Das, Sanchari and Sengupta, Saptarshi},
  booktitle={Annual Conference of the PHM Society},
  volume={16},
  number={1},
  year={2024}
}

@inproceedings{gungor2024roldef,
  title={Roldef: Robust layered defense for intrusion detection against adversarial attacks},
  author={Gungor, Onat and Rosing, Tajana and Aksanli, Baris},
  booktitle={2024 Design, Automation \& Test in Europe Conference \& Exhibition (DATE)},
  pages={1--6},
  year={2024},
  organization={IEEE}
}

@inproceedings{gungor2024rigorous,
  title={Rigorous evaluation of machine learning-based intrusion detection against adversarial attacks},
  author={Gungor, Onat and Li, Elvin and Shang, Zhengli and Guo, Yutong and Chen, Jing and Davis, Johnathan and Rosing, Tajana},
  booktitle={2024 IEEE International Conference on Cyber Security and Resilience (CSR)},
  pages={152--158},
  year={2024},
  organization={IEEE}
}

@inproceedings{kocal2025relate,
  author={Kocal, Cagla Ipek and Gungor, Onat and Tartz, Aaron and Rosing, Tajana and Aksanli, Baris},
  booktitle={2025 IEEE International Conference on Cyber Security and Resilience (CSR)}, 
  title={ReLATE: Resilient Learner Selection for Multivariate Time-Series Classification Against Adversarial Attacks}, 
  year={2025},
  volume={},
  number={},
  pages={419-424},
  doi={10.1109/CSR64739.2025.11130062}}

@misc{individual_household_electric_power_consumption_235,
  author       = {Hebrail, Georges and Berard, Alice},
  title        = {{Individual Household Electric Power Consumption}},
  year         = {2006},
  howpublished = {UCI Machine Learning Repository},
  note         = {{DOI}: https://doi.org/10.24432/C58K54}
}

@article{tang2022survey,
  title={A survey on machine learning models for financial time series forecasting},
  author={Tang, Yajiao and Song, Zhenyu and Zhu, Yulin and Yuan, Huaiyu and Hou, Maozhang and Ji, Junkai and Tang, Cheng and Li, Jianqiang},
  journal={Neurocomputing},
  volume={512},
  pages={363--380},
  year={2022},
  publisher={Elsevier}
}

@inproceedings{tang2021adversarial,
  title={Adversarial attacks to solar power forecast},
  author={Tang, Ningkai and Mao, Shiwen and Nelms, R Mark},
  booktitle={IEEE GLOBECOM},
  pages={1--6},
  year={2021},
  organization={IEEE}
}

@article{morid2023time,
  title={Time series prediction using deep learning methods in healthcare},
  author={Morid, Mohammad Amin and Sheng, Olivia R Liu and Dunbar, Joseph},
  journal={ACM Transactions on Management Information Systems},
  volume={14},
  number={1},
  pages={1--29},
  year={2023},
  publisher={ACM New York, NY}
}

@article{romano2019conformalized,
  title={Conformalized quantile regression},
  author={Romano, Yaniv and Patterson, Evan and Candes, Emmanuel},
  journal={Advances in neural inf. process. systems},
  volume={32},
  year={2019}
}

@inproceedings{chen2019exploiting,
  title={Exploiting vulnerabilities of load forecasting through adversarial attacks},
  author={Chen, Yize and Tan, Yushi and Zhang, Baosen},
  booktitle={Proceedings of the tenth ACM international conference on future energy systems},
  pages={1--11},
  year={2019}
}

@article{szegedyintriguing,
  title={Intriguing properties of neural networks},
  author={Szegedy, Christian and Zaremba, Wojciech and Sutskever, Ilya and Bruna, Joan and Erhan, Dumitru and Goodfellow, Ian and Fergus, Rob}
}

@article{goodfellow2015explaining,
  title={EXPLAINING AND HARNESSING ADVERSARIAL EXAMPLES},
  author={Goodfellow, Ian J and Shlens, Jonathon and Szegedy, Christian},
  journal={stat},
  volume={1050},
  pages={20},
  year={2015}
}

@article{costa2024deep,
  title={How deep learning sees the world: A survey on adversarial attacks \& defenses},
  author={Costa, Joana C and Roxo, Tiago and Proen{\c{c}}a, Hugo and In{\'a}cio, Pedro RM},
  journal={IEEE Access},
  year={2024},
  publisher={IEEE}
}

@article{govindarajulu2023targeted,
  title={Targeted attacks on timeseries forecasting},
  author={Govindarajulu, Yuvaraj and Amballa, Avinash and Kulkarni, Pavan and Parmar, Manojkumar},
  journal={arXiv:2301.11544},
  year={2023}
}

@inproceedings{dang2020adversarial,
  title={Adversarial attacks on probabilistic autoregressive forecasting models},
  author={Dang-Nhu, Rapha{\"e}l and Singh, Gagandeep and Bielik, Pavol and Vechev, Martin},
  booktitle={International Conference on Machine Learning},
  pages={2356--2365},
  year={2020},
  organization={PMLR}
}

@article{wen2023onenet,
  title={Onenet: Enhancing time series forecasting models under concept drift by online ensembling},
  author={Wen, Qingsong and Chen, Weiqi and Sun, Liang and Zhang, Zhang and Wang, Liang and Jin, Rong and Tan, Tieniu and others},
  journal={Advances in Neural Information Processing Systems},
  volume={36},
  pages={69949--69980},
  year={2023}
}

@article{cheng2025adversarial,
  title={Adversarial purification for data-driven power system event classifiers with diffusion models},
  author={Cheng, Yuanbin and Yamashita, Koji and Follum, Jim and Yu, Nanpeng},
  journal={IEEE Transactions on Power Systems},
  year={2025},
  publisher={IEEE}
}

@inproceedings{abdu2020detecting,
  title={Detecting adversarial attacks in time-series data},
  author={Abdu-Aguye, Mubarak G and Gomaa, Walid and Makihara, Yasushi and Yagi, Yasushi},
  booktitle={IEEE ICASSP},
  pages={3092--3096},
  year={2020},
}

@article{ma2018characterizing,
  title={Characterizing adversarial subspaces using local intrinsic dimensionality},
  author={Ma, Xingjun and Li, Bo and Wang, Yisen and Erfani, Sarah M and Wijewickrema, Sudanthi and Schoenebeck, Grant and Song, Dawn and Houle, Michael E and Bailey, James},
  journal={arXiv preprint arXiv:1801.02613},
  year={2018}
}

@incollection{kurakin2018adversarial,
  title={Adversarial examples in the physical world},
  author={Kurakin, Alexey and Goodfellow, Ian J and Bengio, Samy},
  booktitle={Artificial intelligence safety and security},
  pages={99--112},
  year={2018},
  publisher={Chapman and Hall/CRC}
}

@inproceedings{lau2025fast,
  title={Fast and Slow Streams for Online Time Series Forecasting Without Information Leakage},
  author={Lau, Ying-yee Ava and Shao, Zhiwen and Yeung, Dit-Yan},
  booktitle={The Thirteenth International Conference on Learning Representations},
  year={2025}
}

@article{ferreira2023forecasting,
  title={Forecasting network traffic: A survey and tutorial with open-source comparative evaluation},
  author={Ferreira, Gabriel O and Ravazzi, Chiara and Dabbene, Fabrizio and Calafiore, Giuseppe C and Fiore, Marco},
  journal={IEEE Access},
  volume={11},
  pages={6018--6044},
  year={2023},
  publisher={IEEE}
}

@article{wu2022small,
  title={Small perturbations are enough: Adversarial attacks on time series prediction},
  author={Wu, Tao and Wang, Xuechun and Qiao, Shaojie and Xian, Xingping and Liu, Yanbing and Zhang, Liang},
  journal={Information Sciences},
  volume={587},
  pages={794--812},
  year={2022},
  publisher={Elsevier}
}

@inproceedings{gong2019real,
  title={Real-time adversarial attacks},
  author={Gong, Yuan and Li, Boyang and Poellabauer, Christian and Shi, Yiyu},
  booktitle={Proceedings of the 28th International Joint Conference on Artificial Intelligence},
  pages={4672--4680},
  year={2019}
}

@article{lim2021time,
  title={Time-series forecasting with deep learning: a survey},
  author={Lim, Bryan and Zohren, Stefan},
  journal={Philosophical transactions of the royal society a: mathematical, physical and engineering sciences},
  volume={379},
  number={2194},
  year={2021},
  publisher={The Royal Society}
}

@inproceedings{fawaz2019adversarial,
  title={Adversarial attacks on deep neural networks for time series classification},
  author={Fawaz, Hassan Ismail and Forestier, Germain and Weber, Jonathan and Idoumghar, Lhassane and Muller, Pierre-Alain},
  booktitle={IEEE IJCNN},
  pages={1--8},
  year={2019},
}

@article{eren2024comprehensive,
  title={A comprehensive review on deep learning approaches for short-term load forecasting},
  author={Eren, Yavuz and K{\"u}{\c{c}}{\"u}kdemiral, {\.I}brahim},
  journal={Renewable and Sustainable Energy Reviews},
  volume={189},
  pages={114031},
  year={2024},
  publisher={Elsevier}
}

@article{lin2019nesterov,
  title={Nesterov accelerated gradient and scale invariance for adversarial attacks},
  author={Lin, Jiadong and Song, Chuanbiao and He, Kun and Wang, Liwei and Hopcroft, John E},
  journal={arXiv preprint arXiv:1908.06281},
  year={2019}
}

@inproceedings{wang2017time,
  title={Time series classification from scratch with deep neural networks: A strong baseline},
  author={Wang, Zhiguang and Yan, Weizhong and Oates, Tim},
  booktitle={International joint conference on neural networks},
  pages={1578--1585},
  year={2017},
  organization={IEEE}
}

@online{pecanstreet_dataport,
  author  = {{Pecan Street Inc.}},
  title   = {Pecan Street Dataport},
  year    = {2022},
  url     = {https://www.pecanstreet.org/dataport/},
  urldate = {2026-03-08}
}
%\bibliography{biblio}

\end{document}